\documentclass[letterpaper, 10 pt, conference]{ieeeconf}  

\IEEEoverridecommandlockouts                              

\usepackage{bm}
\usepackage{epsfig}
\usepackage{graphicx}
\usepackage{microtype}
\usepackage{xcolor}
\newcommand\hl[1]{%
	\bgroup
	\hskip0pt\color{red!80!black}%
	#1%
	\egroup
}

\title{\LARGE \bf Degeneracy in Self-Calibration Revisited and a Deep Learning Solution for Uncalibrated SLAM}

\author{Bingbing Zhuang$^{1}$ \hfill Quoc-Huy Tran$^{2}$ \hfill Pan Ji$^{2}$ \hfill Gim Hee Lee$^{1}$  \hfill Loong Fah Cheong$^{1}$ \hfill Manmohan Chandraker$^{2,3}$
	\thanks{$^{1}$B. Zhuang, G.H. Lee, and L. F. Cheong are with National University of Singapore {\tt\small zhuang.bingbing@u.nus.edu, dcslgh@nus.edu.sg, eleclf@nus.edu.sg}. \newline$^{2}$Q.-H. Tran, Pan Ji, and M. Chandraker are with NEC Laboratories America, Inc. {\tt\small qhtran@nec-labs.com, manu@nec-labs.com}.\newline$^{3}$M. Chandraker is also with University of California, San Diego {\tt\small mkchandraker@eng.ucsd.edu}.}
}

\begin{document}

\maketitle
\thispagestyle{empty}
\pagestyle{empty}

\begin{abstract}
 Self-calibration of camera intrinsics and radial distortion has a long history of research in the computer vision community. However, it remains rare to see real applications of such techniques to modern Simultaneous Localization And Mapping (SLAM) systems, especially in driving scenarios. In this paper, we revisit the geometric approach to this problem, and provide a theoretical proof that explicitly shows the ambiguity between radial distortion and scene depth when two-view geometry is used to self-calibrate the radial distortion. In view of such geometric degeneracy, we propose a learning approach that trains a convolutional neural network (CNN)  on a large amount of synthetic data. We demonstrate the utility of our proposed method by applying it as a checkerboard-free calibration tool for SLAM, achieving comparable or superior performance to previous learning and hand-crafted methods.
\end{abstract}

\section{Introduction} 

\label{sec:introduction}

With the research efforts in the last decade, SLAM has achieved significant progress and maturity. It has reached the stage that one can expect satisfactory results to be obtained by the various state-of-the-art SLAM systems~\cite{engel2014lsd,mur2015orb,engel2018direct} under well-controlled environments. However, as pointed out in~\cite{yang2018challenges}, many challenges still exist under more unconstrained conditions. These typically include image degradation, e.g., motion blurs~\cite{lee2011simultaneous,park2017joint} and rolling shutter effects~\cite{zhuang2017rolling,Zhuang_2019_CVPR}, and unknown camera calibration including radial distortion. In this paper, we work towards a robust camera calibration system that enables SLAM on a video input without knowing the camera calibration a priori. Specifically, we revisit the ill-posedness of the self-calibration problem as set forth in a conventional geometric approach, and propose an effective data-driven approach. 

Radial distortion estimation has been a long-standing problem in geometric vision. Many minimal solvers (e.g, \cite{kukelova2007minimal,kukelova2015efficient}) utilize two-view epipolar constraints to estimate radial distortion parameters. However, such methods have not been integrated into state-of-the-art SLAM systems~\cite{engel2014lsd,mur2015orb,engel2018direct}. Therefore, its robustness in practical scenarios, e.g. driving scenes~\cite{geiger2012we,dhiman2016continuous,wang2019parametric}, has not been tested as extensively as one would have liked. In this paper, we provide a further theoretical and numerical study on such geometric approaches. Specifically, although it has been mentioned in existing works~\cite{steger2012estimating,wu2014critical} that forward camera motion constitutes a degenerate case for two-view radial distortion self-calibration, we give a more general and formal proof of this observation that explicitly describes the ambiguity between radial distortion and scene depth in two-view geometry. We further show that such degeneracy causes the solution to be very unstable when applied to driving scenes under (near-)forward motion. Moreover, it is well known from previous works~\cite{hartley1993extraction,sturm2001focal} that focal length estimation from two views also suffers from degeneracy under various configurations, e.g., when the optical axes of the two views intersect at a single point.

\begin{figure}[!!t] 
	
	\begin{center} 
		
		\includegraphics[width=1.0\linewidth, trim = 0mm 0mm 0mm 0mm, clip]{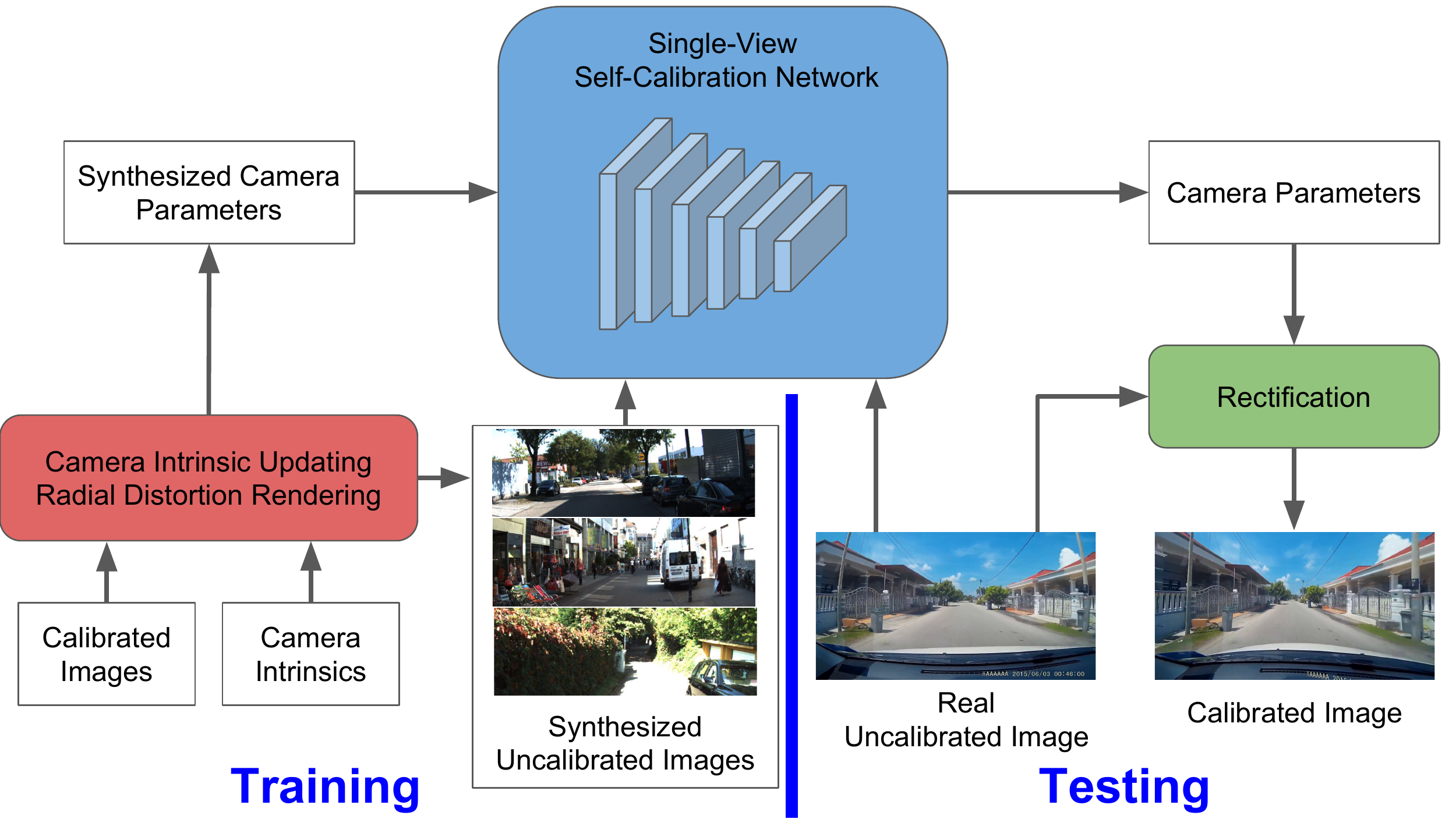} 
		
	\end{center} 
	\vspace{-0.4cm}
	\caption{Overview of our approach to single-view camera self-calibration. From a set of calibrated images with known camera intrinsics and no radial distortion, we generate synthetic uncalibrated images for training our network by randomly adjusting camera intrinsics and randomly adding radial distortion. At testing, given a single real uncalibrated image, our network predicts accurate camera parameters, which are used for producing the calibrated image.} 
	
	\label{fig:selfcalibration} 
	
\end{figure}

To mitigate such geometric degeneracy, we propose to exploit a data-driven approach that learns camera self-calibration from a single image (see Fig.~\ref{fig:selfcalibration}). This allows us to leverage any regularity in the scenes so that the intrinsic degeneracies of the self-calibration problem can be overcome. Specifically, we target at driving scenes and synthesize a large amount of images with different values of radial distortion, focal length, and principal point. We show that a network trained on such synthetic data is able to generalize well to unseen real data. Furthermore, in this work we exploit the recent success of CNNs with deep supervision~\cite{lee2015deeply,li2017deep,lee2017roomnet,li2018deep,fathy2018hierarchical}, which suggests that providing explicit supervisions to intermediate layers contributes to improved regularization. We observe that it performs better than a typical multi-task network~\cite{caruana1997multitask} in terms of radial distortion and focal length estimation. Lastly, we empirically show that it is feasible to perform SLAM on uncalibrated videos from KITTI Raw~\cite{geiger2012we} and YouTube with the self-calibration by our network.

\noindent In summary, our contributions include: 

\begin{itemize} 	
	\item Using a general distortion model and a discrete Structure from Motion (SfM) formulation, we prove that two-view self-calibration of radial distortion is degenerate under forward motion. This is important for understanding intrinsic and algorithm-independent properties of radially-distorted two-view geometry estimation. 	
	\item We propose a CNN-based approach which leverages deep supervision for learning both radial distortion and camera intrinsics from a single image. We demonstrate the application of our single-view camera self-calibration network to SLAM operating on uncalibrated videos from KITTI Raw and YouTube.  	
\end{itemize} 

\section{Related Work}
\label{sec:relatedwork}

\noindent \textbf{Geometric Degeneracy With Unknown Radial Distortion:} It has been shown that self-calibration of unknown radial distortion parameters using two-view geometry suffers from degeneracy under forward motion. Specifically, \cite{steger2012estimating} shows that two-view radially-distorted geometry under forward motion is degenerate with the division distortion model~\cite{fitzgibbon2001simultaneous} and a discrete SfM formulation. More recently, critical configurations of radially-distorted cameras under infinitesimal motions are identified and proved in~\cite{wu2014critical} for a general distortion model, but with an approximate differential SfM formulation~\cite{horn1988motion}. In this work, we prove that radially-distorted two-view geometry under forward motion is degenerate using a more general setting than~\cite{steger2012estimating,wu2014critical}, i.e. a general distortion model and a discrete SfM formulation.

\vspace{0.2cm}
\noindent \textbf{Multiple-View Camera Self-Calibration: }Multiple-view methods use multiple-view geometry (usually two-view epipolar constraints) in radially-distorted images to correct radial distortion~\cite{kukelova2007minimal,byrod2008fast,kukelova2015efficient} as well as estimate camera intrinsics~\cite{jiang2014minimal}, via keypoint correspondences between the images. Despite being able to handle different camera motions or scene structures, they need two input images and often require solving complex polynomial systems.

\vspace{0.2cm}
\noindent \textbf{Single-View Camera Self-Calibration: }Single-view methods rely on extracted line/curve features in the input image to remove radial distortion~\cite{bukhari2013automatic,mei2015radial} and/or compute camera intrinsics~\cite{wildenauer2013closed,zhang2015line,santana2017automatic,antunes2017unsupervised}. Moreover, some of them assume special scene structures, e.g., Manhattan world~\cite{antunes2017unsupervised}. One problem with these hand-crafted methods is that they cannot work well when the scenes contain very few line/curve features or the underlying assumptions on the scene structures are violated. Recently, single-view CNN-based methods have been proposed for radial distortion correction~\cite{rong2016radial} and/or camera intrinsic calibration~\cite{yin2018fisheyerecnet,bogdan2018deepcalib}. Our method also uses powerful CNN-extracted features, however, unlike~\cite{yin2018fisheyerecnet,bogdan2018deepcalib} which use multi-task supervision, we apply deep supervision. More importantly, we demonstrate the practical utility of the deep learning method for uncalibrated SLAM.

\section{Degeneracy in Two-View Radial Distortion Self-Calibration under Forward Motion}
\label{sec:degeneracy}

\begin{figure}[!!t]
	\begin{center}
		\includegraphics[width=0.8\linewidth, trim = 0mm 15mm 65mm 0mm, clip]{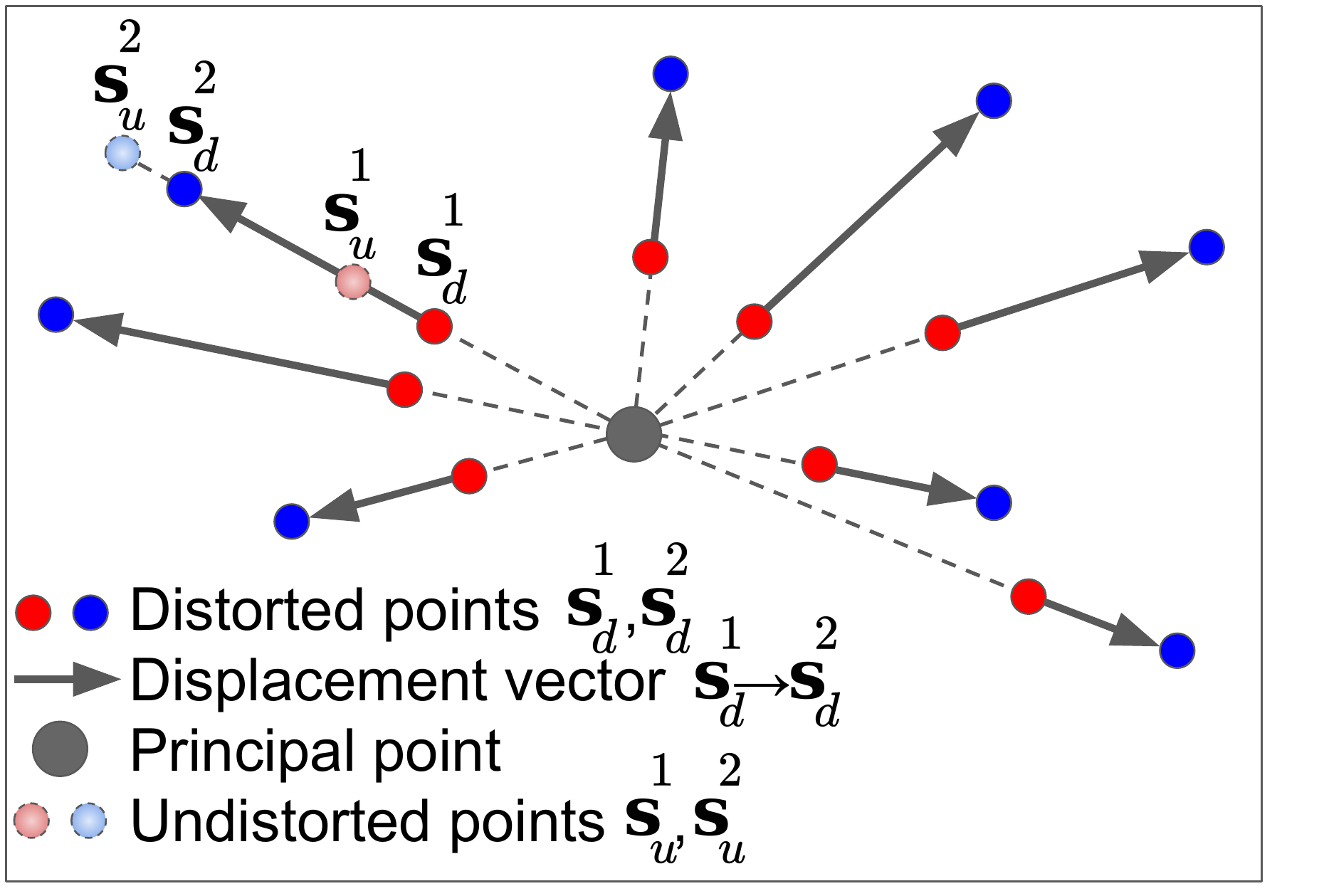}
	\end{center}
	\vspace{-0.4cm}
	\caption{Degeneracy in two-view radial distortion self-calibration under forward motion. There are infinite number of valid combinations of radial distortion and scene structure, including the special case with zero radial distortion.}
	\label{fig:degeneracy}
\end{figure} 

We denote the 2D coordinates of a distorted point on a normalized image plane as $\bm{s}_d = [x_d,y_d]^\top$ and the corresponding undistorted point as $\bm{s}_u = [x_u,y_u]^\top = f(\bm{s}_d;\bm{\theta})\bm{s}_d$.$\bm{\theta}$ is the radial distortion parameters and $f(\bm{s}_d;\bm{\theta})$ is the undistortion function which scales $\bm{s}_d$ to $\bm{s}_u$. The specific form of $f(\bm{s}_d;\bm{\theta})$ depends on the radial distortion model being used. For instance, we have $f(\bm{s}_d;\lambda) = \frac{1}{1+\lambda r^2}$ for the division model~\cite{fitzgibbon2001simultaneous} with one parameter, or we have $f(\bm{s}_d;\lambda) = 1+\lambda r^2$ for the polynomial model~\cite{wang2008new} with one parameter. In both models, $\lambda$ is the 1D radial distortion parameter and $r = \sqrt{x_d^2 + y_d^2}$ is the distance from the principal point. We use the general form $f(\bm{s}_d;\bm{\theta})$ for the analysis below.

We formulate the two-view geometric relationship under forward motion, i.e., how a pure translational camera motion along the optical axis is related to the 2D correspondences and their depths. Let us consider a 3D point $\bm{S}$, expressed as $\bm{S}_1 = [X_1,Y_1,Z_1]^\top$ and $\bm{S}_2 = [X_2,Y_2,Z_2]^\top$, respectively, in the two camera coordinates. Under forward motion, we have $\bm{S}_2 = \bm{S}_1-\bm{T}$ with $\bm{T} = [0,0,t_z]^\top$. Without loss of generality, we fix $t_z = 1$ to remove the global scale ambiguity. Projecting the above relationship onto the image planes, we obtain  $\bm{s}_u^2 = \frac{Z_1}{Z_1-1}\bm{s}_u^1$, where $\bm{s}_u^1$ and $\bm{s}_u^2$ are the 2D projections of $\bm{S}_1$ and $\bm{S}_2$, respectively (i.e., $\lbrace \bm{s}_u^1, \bm{s}_u^2 \rbrace$ is a 2D correspondence). Expressing the above in terms of the observed distorted points $\bm{s}_d^1$ and $\bm{s}_d^2$ yields:
\begin{equation}
f(\bm{s}_d^2;\bm{\theta}_2)\bm{s}_d^2 = \frac{Z_1}{Z_1-1}f(\bm{s}_d^1;\bm{\theta}_1)\bm{s}_d^1,
\label{eq:degeneracy}
\end{equation}
where $\bm{\theta}_1$ and $\bm{\theta}_2$ represent radial distortion parameters in the two images respectively (note that $\bm{\theta}_1$ may differ from $\bm{\theta}_2$).

Eq.~\ref{eq:degeneracy} represents all the information available for estimating the radial distortion and the scene structure. 
However, the correct radial distortion and point depth cannot be determined from the above equation. We can replace the ground truth radial distortion denoted by $\lbrace \bm{\theta}_1, \bm{\theta}_2 \rbrace$
with a fake radial distortion $\lbrace \bm{\theta}_1^\prime, \bm{\theta}_2^\prime \rbrace$ and the ground truth point depth $Z_1$ for each 2D correspondence with the following fake depth $Z_1^\prime$ such that Eq.~\ref{eq:degeneracy} still holds:
\begin{equation}
Z_1^\prime = \frac{\alpha Z_1}{(\alpha-1)Z_1+1},~~~\alpha = \frac{f(\bm{s}_d^2;\bm{\theta}_2^\prime)f(\bm{s}_d^1;\bm{\theta}_1)}{f(\bm{s}_d^1;\bm{\theta}_1^\prime)f(\bm{s}_d^2;\bm{\theta}_2)}.
\label{eq:fakeZ}
\end{equation}
In particular, we can possibly set $\forall \bm{s}_d^1: f(\bm{s}_d^1;\bm{\theta}_1^\prime) = 1$, $\forall \bm{s}_d^2: f(\bm{s}_d^2;\bm{\theta}_2^\prime) = 1$ as the fake radial distortion, and use the corrupted depth $Z_1^\prime$ computed according to Eq.~\ref{eq:fakeZ} so that Eq.~\ref{eq:degeneracy} still holds. This special solution corresponds to  the pinhole camera model, i.e., $\bm{s}_u^1 = \bm{s}_d^1$ and $\bm{s}_u^2 = \bm{s}_d^2$. In fact, this special case can be inferred more intuitively. Eq.~\ref{eq:degeneracy} indicates that all 2D points move along 2D lines radiating from the principal point, as illustrated in Fig.~\ref{fig:degeneracy}. This pattern is exactly the same as in the pinhole camera model, and is the sole cue to recognize the forward motion.

Intuitively, the 2D point movements induced by radial distortion alone, e.g., between $\bm{s}_u^1$ and $\bm{s}_d^1$, or between $\bm{s}_u^2$ and $\bm{s}_d^2$, are along the same direction as the 2D point movements induced by forward motion alone, e.g., between $\bm{s}_u^1$ and $\bm{s}_u^2$ (see Fig.~\ref{fig:degeneracy}). 
Hence, radial distortion only affects the magnitudes of 2D point displacements but not their directions in cases of forward motion. Furthermore, such radial distortion can be compensated with an appropriate corruption in the depths so that a corrupted scene structure that explains the image observations, i.e., 2D correspondences, exactly in terms of reprojection errors can still be recovered.

We arrive at the following proposition:

\vspace{0.2cm}
\noindent \textbf{Proposition: }\textit{Two-view radial distortion self-calibration is degenerate for the case of pure forward motion. In particular, there are infinite number of valid combinations of radial distortion and scene structure, including the special case of zero radial distortion.}

\section{Learning Radial Distortion and Camera Intrinsics from a Single Image}
\label{sec:approach}

In this section, we present the details of our learning approach for single-view camera self-calibration. Fig.~\ref{fig:selfcalibration} shows an overview of our approach.

\subsection{Network Architecture}
\label{sec:architecture}

We adopt the ResNet-34 architecture~\cite{he2016deep} as our base architecture and make the following changes. We first remove the last average pooling layer which was designed for ImageNet classification. Next, we add a set of $3 \times 3$ convolutional layers (each followed by a BatchNorm layer and a ReLU activation layer) for extracting features, and then a few $1 \times 1$ convolutional layers (with no bias) for regressing the output parameters, including radial distortion $\lambda$ (defined on the normalized plane), focal length $f$, and principal point $\lbrace c_x, c_y \rbrace$. Here, we adopt the widely used division model~\cite{fitzgibbon2001simultaneous} with one parameter for radial distortion (the corresponding undistortion only requires to solve a simple quadratic equation).

\begin{figure}[!!t]
	\begin{center}
		\includegraphics[width=1.0\linewidth, trim = 0mm 0mm 50mm 60mm, clip]{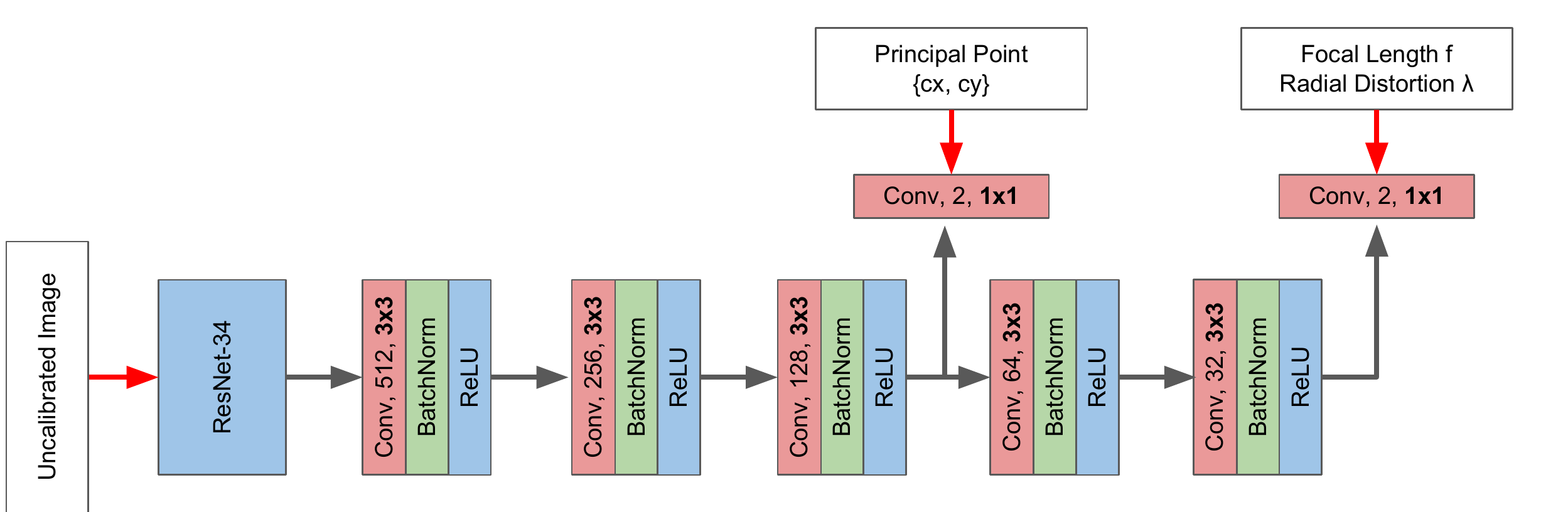}
	\end{center}
	\vspace{-0.4cm}
	\caption{Detailed architecture of our single-view camera self-calibration network. White boxes denote input data (left) or supervision signals (top).}
	\label{fig:architecture}
\end{figure}

To apply deep supervision~\cite{li2018deep} on our problem, we need to select a dependence order between the predicted parameters. 
Knowing that: (1) a known principal point is clearly a prerequisite for estimating radial distortion, and (2) image appearance is affected by the composite effect of radial distortion and focal length, we predict the parameters in the following order: (1) principal point in the first branch, and (2) both focal length and radial distortion in the second branch. Fig.~\ref{fig:architecture} shows our network architecture. We have also tried separating the prediction of $\lambda$ and $f$ in two branches but it does not perform as well. We train our network by using an $L_1$-norm regression loss for each predicted parameter ($L_1$-norm is preferred since it is more robust than $L_2$-norm).

\subsection{Training Data Generation}
\label{sec:datagen}

Since there is no large-scale dataset of real uncalibrated images with ground truth camera parameters available, we  generate synthetic images with ground truth parameters based on the KITTI Raw dataset~\cite{geiger2012we} to train our network. We use all sequences in the `City' and `Residential' categories, except for 5 sequences that contain mostly static scenes (i.e., sequences `0017', `0018', `0057', `0060' in `2011\_09\_26', and sequence `0001' in `2011\_09\_28') and 2 sequences that are used for validation (i.e., sequence `0071' in `2011\_09\_29') and evaluation in Sec.~\ref{sec:experiments} (i.e., sequence `0027' in `2011\_10\_03'). In total, we use 42 sequences with around 30K frames to generate the training data.

For each calibrated image with known camera intrinsics and no radial distortion, we randomly vary the camera intrinsics ($\lbrace c_x, c_y \rbrace \in [0.45,0.55] \times [0.45,0.55]$, $f \in [0.6,1.8]$) and randomly add a radial distortion component ($\lambda \in [-1.0,0]$) --- note that $f$ and $c_x$ are normalized by image width, while $c_y$ is normalized by image height. We find the above parameter ranges to be sufficient for the videos we test as shown in Sec.~\ref{sec:experiments}. The modified parameters are then used to generate a new uncalibrated image accordingly by generating radial distortion effect, image cropping, and image resizing. For each calibrated image, we repeat the above procedure 10 times to generate 10 uncalibrated images, resulting in a total of around 300K synthetic uncalibrated images with known ground truth camera parameters for training our network.

\subsection{Training Details}
\label{sec:traindetail}

We set the weight of each regression loss to 1.0 and use the ADAM optimization~\cite{kingma2014adam} with learning rate of $10^{-3}$ to train our network. We use the pre-trained weights of ResNet-34 on ImageNet classification to initialize the common layers in our network, and the newly added layers are randomly initialized by using~\cite{glorot2010understanding}. The input image size is $320 \times 960$ pixels. We set the batch size to 40 images and train our network for 6 epochs (chosen based on validation performance). Our network is implemented in PyTorch~\cite{paszke2017automatic}.

\section{Experiments}
\label{sec:experiments}

We first analyze the performance of traditional geometric method and our learning method for radial distortion self-calibration in near-degenerate cases in Sec.~\ref{sec:degen_exp}. Next, we compare our method against state-of-the-art methods for single-view camera self-calibration in Sec.~\ref{sec:calib_exp}. Lastly, Sec.~\ref{sec:slam_exp} demonstrates the application of our method to uncalibrated SLAM. Note that for our method, we use our model trained solely on the synthetic data from Sec.~\ref{sec:datagen} without any fine-tuning with real data.

\vspace{0.2cm}
\noindent \textbf{Competing Methods: }We benchmark our method (`DeepSup') against state-of-the-art methods~\cite{wildenauer2013closed,santana2017automatic,bogdan2018deepcalib} for single-view camera self-calibration. Inspired by~\cite{bogdan2018deepcalib}, we add a multi-task baseline (`MultiTask'), which has similar network architecture as ours but employing multi-task supervision, i.e., predicting principal point, focal length, and radial distortion all at the final layer. We train both `MultiTask' and `DeepSup' on the same set of synthetic data in Sec.~\ref{sec:datagen}. In addition, we evaluate the performance of two hand-crafted methods: (1) 'HandCrafted', which uses lines or curves and Hough Transformation to estimate radial distortion and principal point \cite{santana2017automatic} and  (2) 'HandCrafted+OVP', which uses orthogonal vanishing points to estimate radial distortion and focal length \cite{wildenauer2013closed}. For `HandCrafted+OVP', we use the binary code provided by the authors. For `HandCrafted', we need to manually upload each image separately to the web interface\footnote{http://dev.ipol.im/$\sim$asalgado/ipol\_demo/workshop\_perspective/} to obtain the results, thus we only show a few qualitative comparisons with this method in the following experiments. 

\vspace{0.2cm}
\noindent \textbf{Test Data: }Our test data consists of 4 real uncalibrated sequences. In particular, 1 sequence is originally provided by KITTI Raw~\cite{geiger2012we} (i.e., sequence `0027' in `2011\_10\_03' --- with over 4.5K frames in total and no overlap with the sequences used for generating the training data in Sec.~\ref{sec:datagen}) and 3 sequences are extracted from YouTube videos (with 1.2K, 2.4K, and 1.6K frames respectively). Ground truth camera parameters and camera poses are available for the KITTI Raw test sequence (we convert the distortion model used in KITTI Raw, i.e., OpenCV calibration toolbox, to the division model~\cite{fitzgibbon2001simultaneous} with one parameter by regressing the new model parameter). Ground truth for YouTube test sequences is not available. We also note that the aspect ratio of the YouTube images is different from that of the training data. Therefore, we first crop the YouTube images to match the aspect ratio of the training data before feeding them to the network, and then convert the predicted camera parameters according to the aspect ratio of the original YouTube images.


\subsection{Geometric Approach vs. Learning Approach}
\label{sec:degen_exp}

\begin{figure}[!!t]
	\begin{center}
		\includegraphics[width=1.0\linewidth, trim = 10mm 0mm 10mm 0mm, clip]{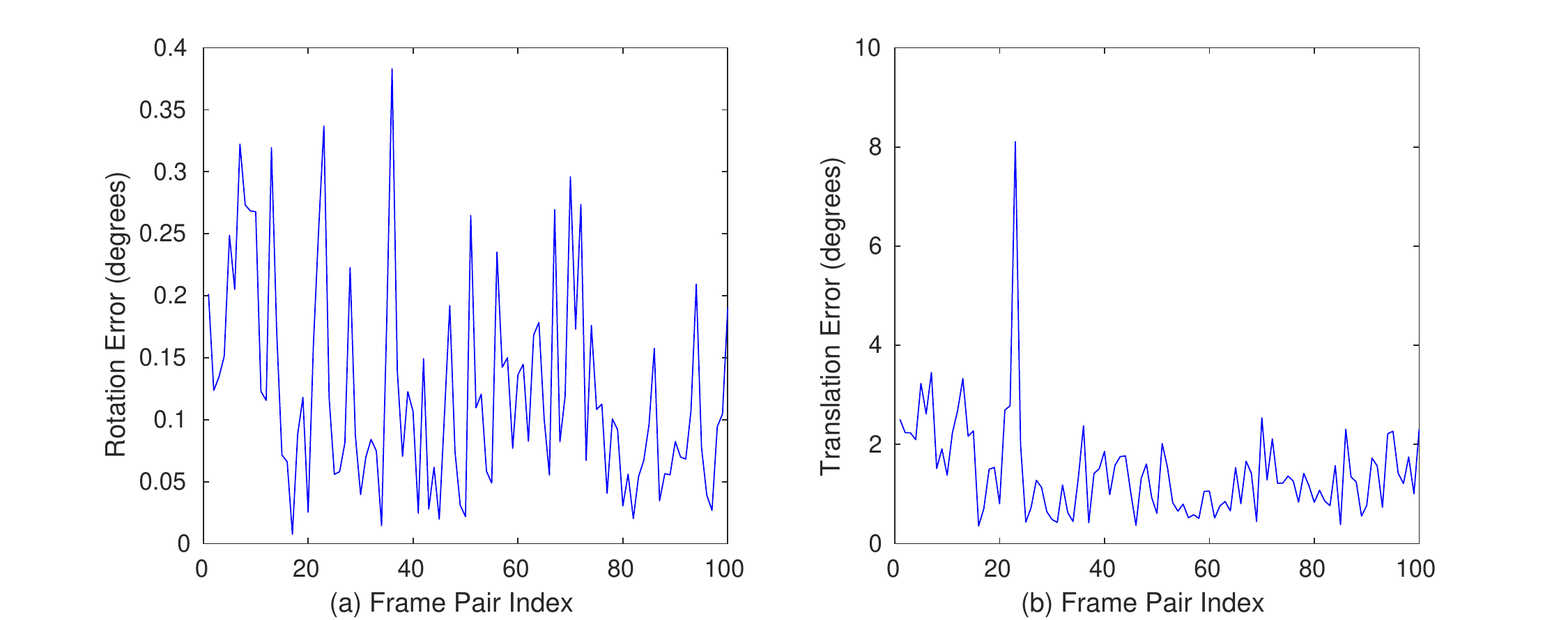}
	\end{center}
	\vspace{-0.4cm}
	\caption{Relative pose estimation by geometric approach. (a) Rotation errors. (b) Translation errors.  }
	\label{fig:degen_pose}
\end{figure}

\begin{figure}[!!t]
	\begin{center}
		\includegraphics[width=1.0\linewidth, trim = 10mm 0mm 10mm 0mm, clip]{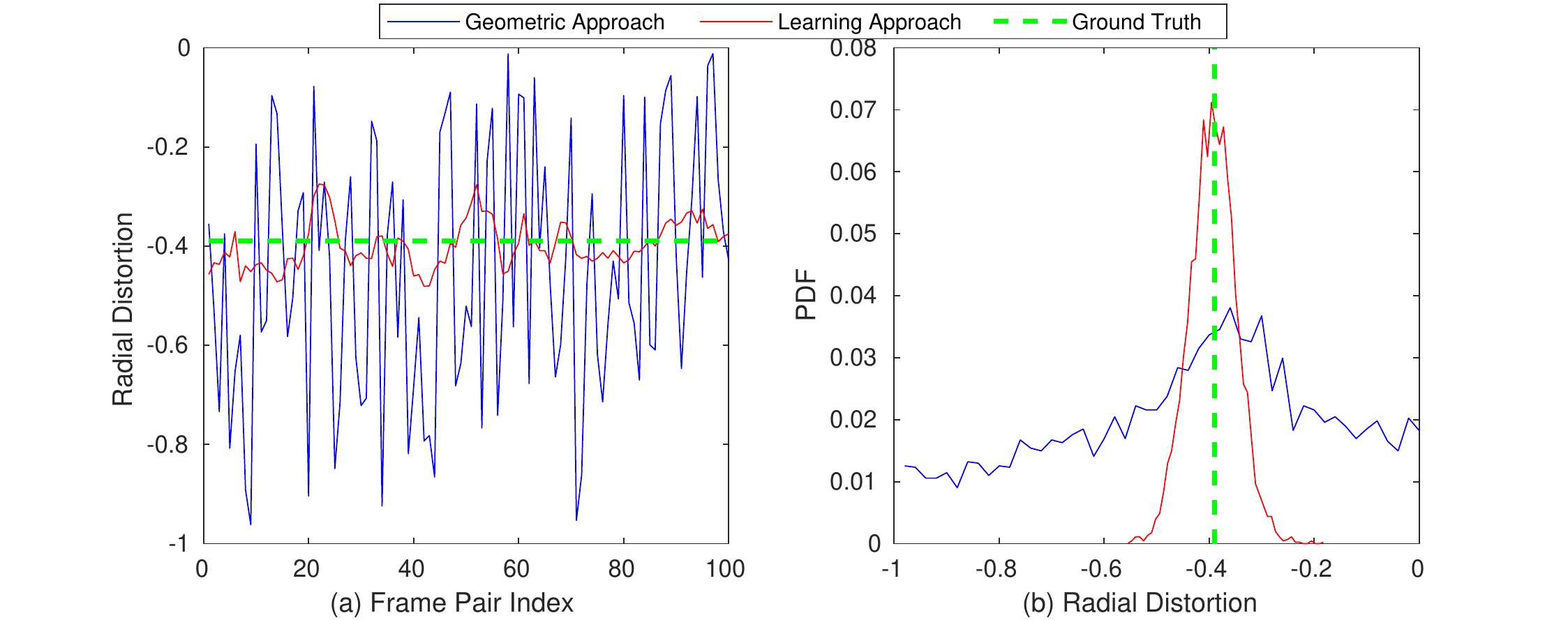}
	\end{center}
	\vspace{-0.4cm}
	\caption{Radial distortion estimation by geometric and learning approaches. (a) Radial distortion estimates. (b) PDF of radial distortion estimates.}
	\label{fig:degen_radial}
\end{figure}

\begin{figure}[!!t]
	\begin{center}
		\includegraphics[width=1.0\linewidth, trim = 5mm 0mm 10mm 0mm, clip]{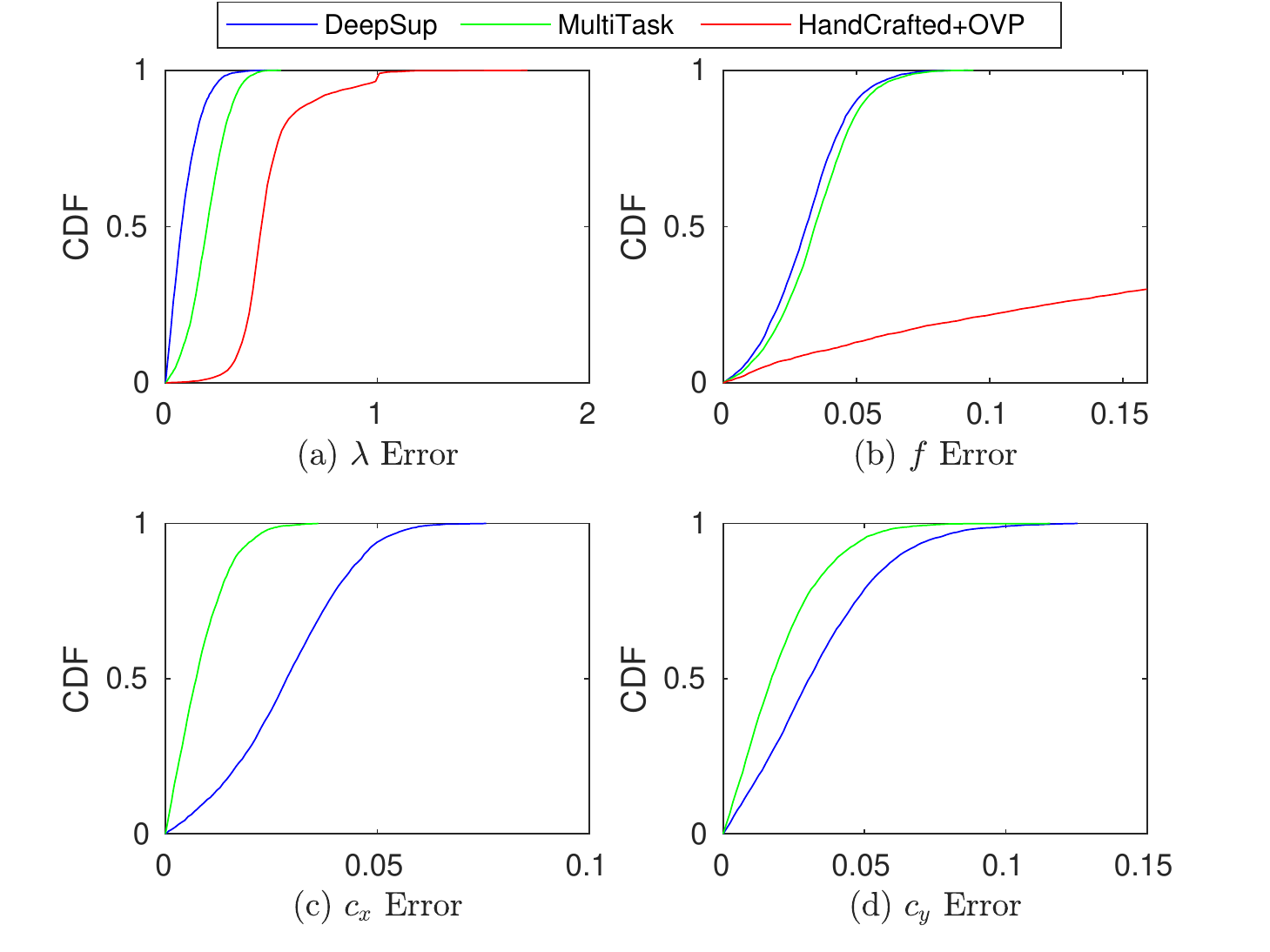}
	\end{center}
	\vspace{-0.4cm}
	\caption{Quantitative results of single-view camera self-calibration on the KITTI Raw test sequence. (a) $\lambda$ errors. (b) $f$ errors. (c) $c_x$ errors. (d) $c_y$ errors. In (b), we only show the CDF corresponding to the smallest 30\% $f$ errors for visibility. `HandCrafted+OVP' assumes known principal point and hence is excluded in (c)-(d).}
	\label{fig:kitti_quantitative}
\end{figure}

\begin{figure*}[!!t]
	\begin{center}
		\includegraphics[width=1.0\linewidth, trim = 0mm 20mm 260mm 0mm, clip]{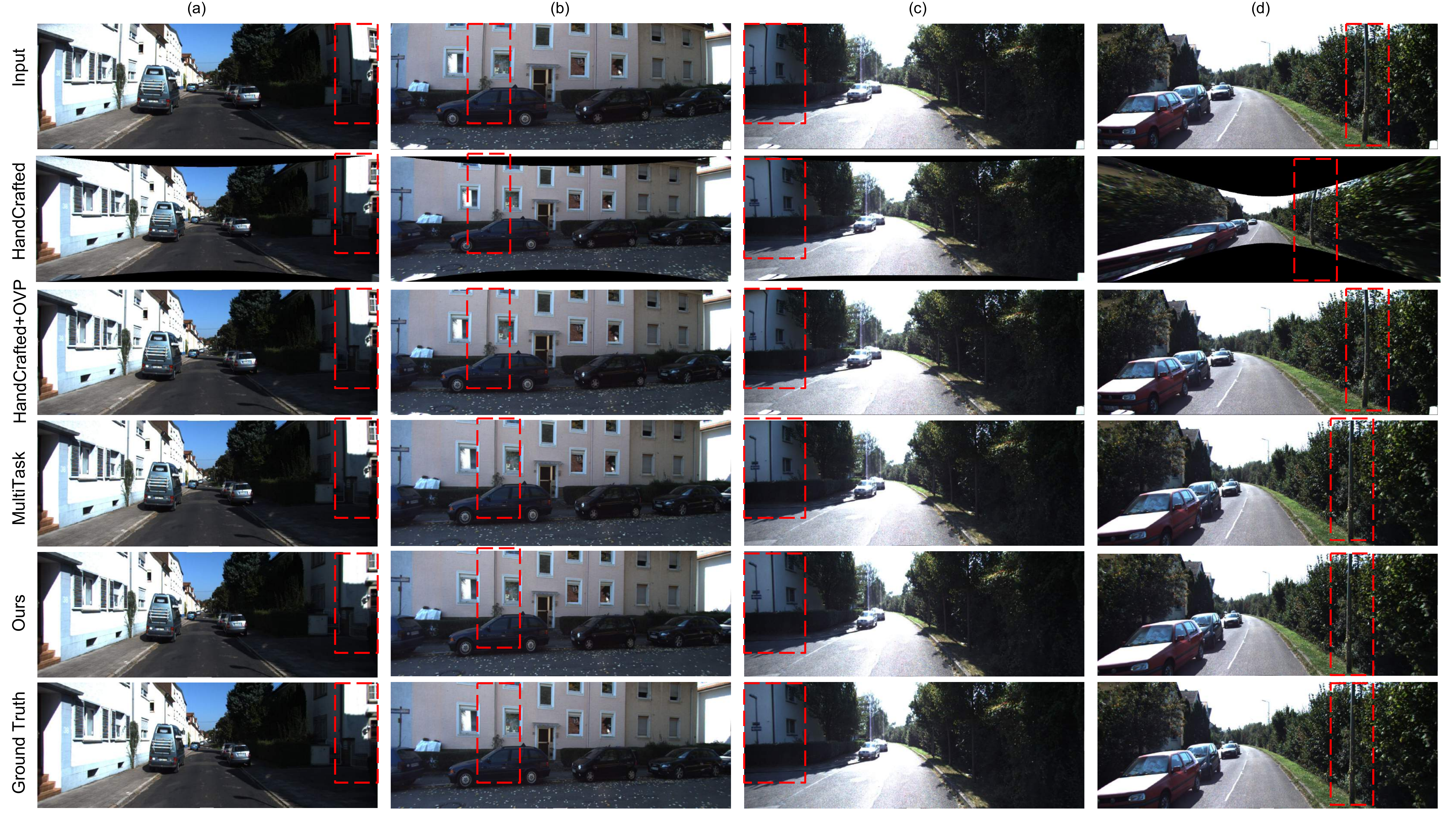}
	\end{center}
	\vspace{-0.4cm}
	\caption{Qualitative results of single-view camera self-calibration on the KITTI Raw test sequence.}
	\label{fig:kitti_qualitative}
\end{figure*}

We have shown in Sec.~\ref{sec:degeneracy} that estimating radial distortion from radially-distorted two-view geometry is degenerate when the camera undergoes ideal forward motion. Here, we investigate the behavior of traditional geometric approach in a practical scenario, i.e., the KITTI Raw test sequence, where the camera motion is near-forward motion. To this end, we implement a minimal solver (9-point algorithm) to solve for both radial distortion $\lambda$ (i.e., division model~\cite{fitzgibbon2001simultaneous} with one parameter) and relative pose by using the following radial distortion aware epipolar constraint: 
\begin{equation}
[x_d^2,~y_d^2,~1+\lambda(x_d^2+y_d^2)]~\bm{E}~[x_d^2,~y_d^2,~1+\lambda(x_d^2+y_d^2)]^\top = 0, 
\end{equation}
where $[x_d^{\{1,2\}},~y_d^{\{1,2\}}]^\top$ is the observed 2D correspondence in the two images and $\bm{E}$ is the essential matrix where the relative pose can be extracted. We apply the minimal solver within RANSAC~\cite{fischler1981paradigm,tran2014robust} to the first 100 consecutive frame pairs of the KITTI Raw test sequence. We first plot the errors of relative pose estimation by the above geometric approach in Fig.~\ref{fig:degen_pose}, which shows that both rotation and translation are estimated with relatively small errors. Next, we plot the results of radial distortion estimation by the geometric approach in Fig.~\ref{fig:degen_radial}, including the radial distortion estimates in (a) and the probability density function (PDF) of the estimates in (b). In addition, we include the results of our learning approach in Fig.~\ref{fig:degen_radial} for comparison. Note that for a fair comparison, in the geometric approach we exclude any minimal sample yielding radial distortion estimate outside $[-1.0,0]$ (which is the radial distortion range of the training data for our learning approach). From the results, it is evident that compared to the geometric approach, the radial distortions estimated from our learning approach are more concentrated around the ground truth ($\lambda = -0.39$), indicating a more robust performance in the (near-)degenerate setting.

\subsection{Single-View Camera Self-Calibration}
\label{sec:calib_exp}

\noindent\textbf{KITTI Raw: }We first quantitatively evaluate the estimation accuracy of radial distortion $\lambda$, focal length $f$, and principal point {$c_x$, $c_y$} against their ground truth values of $\bar{\lambda}$, $\bar{f}$, $\bar{c_x}$, $\bar{c_y}$. We compute the relative errors $\frac{|\lambda-\bar{\lambda}|}{|\bar{\lambda}|}$, $\frac{|f-\bar{f}|}{|\bar{f}|}$, $\frac{|c_x-\bar{c_x}|}{|\bar{c_x}|}$, $\frac{|c_y-\bar{c_y}|}{|\bar{c_y}|}$ and plot the cumulative distribution function (CDF) of the errors from all test images in Fig.~\ref{fig:kitti_quantitative}. It is clear that CNN-based methods (i.e., `DeepSup' and `MultiTask') outperform the hand-crafted method ~\cite{wildenauer2013closed} (i.e., `HandCrafted+OVP') by a significant margin. In particular, we find that the focal lengths estimated by `HandCrafted+OVP' are evidently unstable and suffer from large errors. We also observe that `DeepSup' predicts $\lambda$ and $f$ more accurately than `MultiTask', although it has lower accuracy on $c_x$ and $c_y$ estimation than `MultiTask'.

Next, we show a few qualitative results in Fig.~\ref{fig:kitti_qualitative}. We can see in (a)-(b) that `HandCrafted+OVP' and `HandCrafted' are able to produce reasonable results because the scenes in (a)-(b) contain many structures with line/curve features. However, when the scenes become more cluttered or structure-less as in (c)-(d), such traditional methods cannot work well, especially, the hand-crafted method~\cite{santana2017automatic} (i.e., `HandCrafted') produces significantly distorted result in (d). In contrast, `DeepSup' and `MultiTask' achieve good performance, even in the challenging cases of cluttered or structure-less scenes in (c)-(d). Compared to `MultiTask', `DeepSup' produces results that are visually closer to ground truth. For instance, the line is more straight in (a) and the perspective effect is kept better in (c) for `DeepSup'.

\begin{figure*}[!!t]
	\begin{center}
		\includegraphics[width=1.0\linewidth, trim = 0mm 15mm 65mm 0mm, clip]{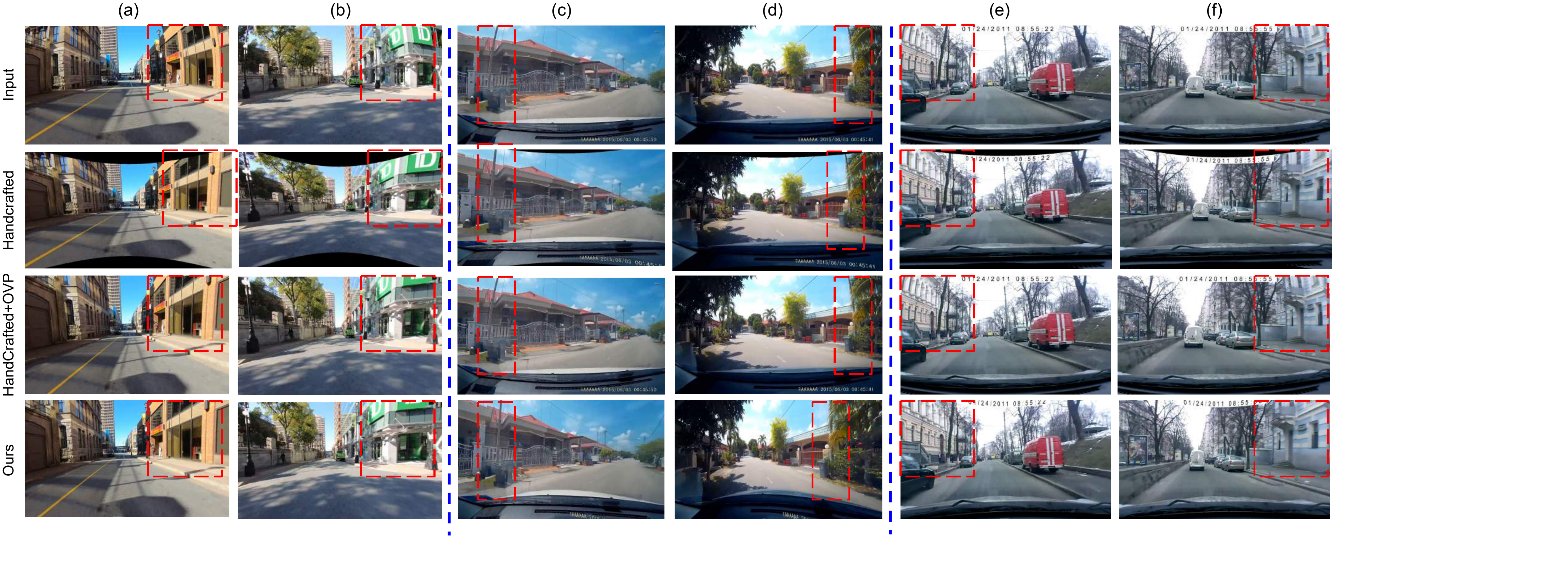}
	\end{center}
	\vspace{-0.4cm}
	\caption{Qualitative results of single-view camera self-calibration on YouTube test sequences. (a)-(b) Sequence 1. (c)-(d) Sequence 2. (e)-(f) Sequence 3.}
	\label{fig:youtube_qualitative}
\end{figure*}

\vspace{0.2cm}
\noindent \textbf{YouTube: }Due to lack of ground truth, we only show a few qualitative results for YouTube test sequences in Fig.~\ref{fig:youtube_qualitative}. From the results, `DeepSup' is able to perform relatively well for all 3 sequences. It is worth noting that the cameras in Sequences 2 and 3 are placed inside the car, thus the hood of the car always appears in the images. Although the hood of the car is not in our synthetic training data in Sec.~\ref{sec:datagen}, our network is able to deal with it and extract useful information from other regions. `MultiTask' produces similar results, which are not shown in this figure. In addition, we observe that although `HandCrafted+OVP' works reasonably for Sequence 1, which mainly consists of well-structured scenes, it fails to calibrate Sequences 2 and 3, where the scenes are more cluttered or structure-less. Similarly, `HandCrafted' performs reasonably for Sequences 1 and 3, but fails for Sequence 2.

\subsection{Uncalibrated SLAM}
\label{sec:slam_exp}

\begin{figure}[!!t]
	\begin{center}
		\includegraphics[width=1.0\linewidth, trim = 0mm 0mm 95mm 50mm, clip]{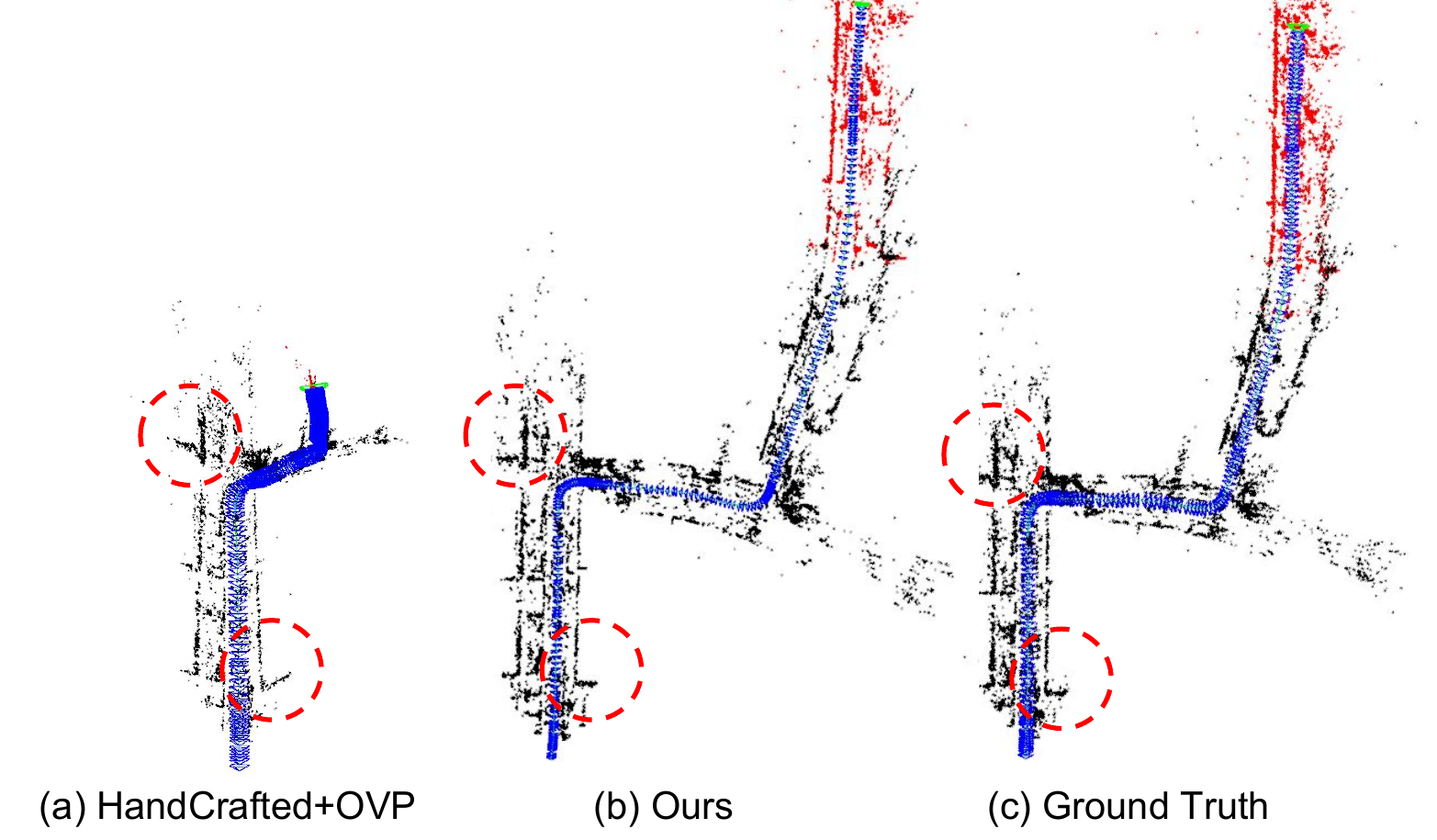}
	\end{center}
	\vspace{-0.4cm}
	\caption{Qualitative results of uncalibrated SLAM on the first 400 frames of the KITTI Raw test sequence. Red points denote the latest local map.}
	\label{fig:kitti_400}
\end{figure}

\begin{figure}[!!t]
	\begin{center}
		\includegraphics[width=1.0\linewidth, trim = 0mm 0mm 55mm 20mm, clip]{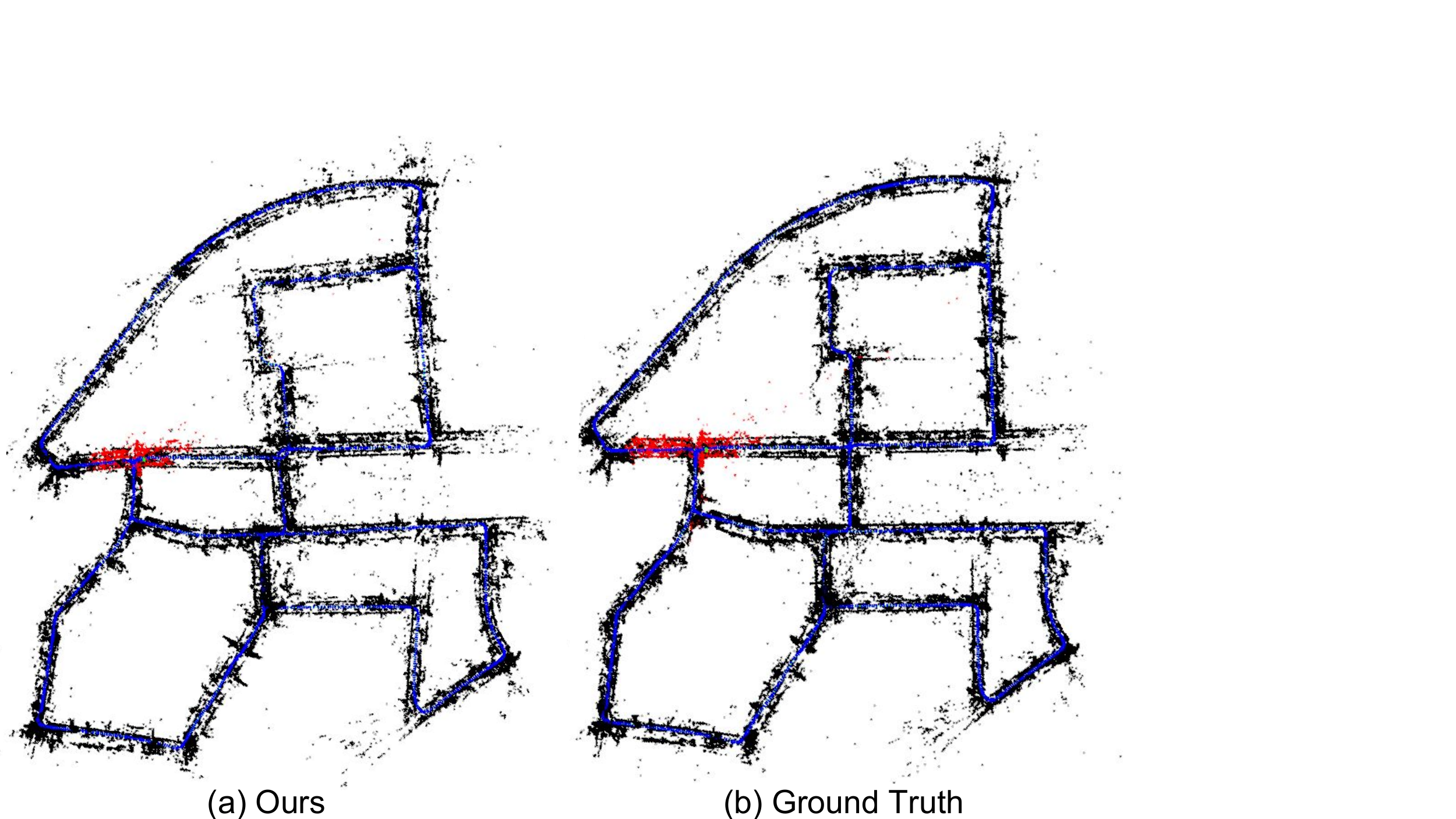}
	\end{center}
	\vspace{-0.4cm}
	\caption{Qualitative results of uncalibrated SLAM on the entire KITTI Raw test sequence. Red points denote the latest local map.}
	\label{fig:kitti_all}
\end{figure}

\begin{table}[t]
\centering
\begin{tabular}{| c | c |}
\hline
Methods & RMSE (m) \\
\hline
HandCrafted+OVP & X \\
Ours 		    & 8.04 \\
Ground Truth    & 7.29 \\
\hline
\end{tabular}
\caption{Quantitative comparisons in terms of median RMSE of keyframe trajectory over 5 runs on the entire KITTI Raw test sequence. X denotes that `HandCrafted+OVP' breaks ORB-SLAM in all 5 runs.}
\label{tab:rmse}
\end{table}

We now demonstrate the application of our single-view camera self-calibration method to SLAM with uncalibrated videos, where calibration with a checkerboard pattern is not available or prohibited. For this purpose, we introduce a two-step approach. We first apply our method on the first 100 frames of the test sequence and take the median values of the per-frame outputs as the estimated  parameters, which are then used to calibrate the entire test sequence. Next, we employ state-of-the-art SLAM systems designed for calibrated environments (here we use ORB-SLAM~\cite{mur2015orb}) on the calibrated images. We compare our two-step approach against similar two-step approaches which use `HandCrafted+OVP' or ground truth parameters (termed as `Ground Truth' --- only available for the KITTI Raw test sequence) for calibration.

\vspace{0.2cm}
\noindent \textbf{KITTI Raw: }We first evaluate the performance of the above approaches on a subset of the KITTI Raw test sequence (i.e., the first 400 frames). Fig.~\ref{fig:kitti_400} presents qualitative results. From the results, although `HandCrafted+OVP' does not break ORB-SLAM in the initial forward motion stage, the reconstructed scene structure and camera trajectory are distorted evidently (see red circles). Subsequently, ORB-SLAM breaks down when the car turns. In contrast, the scene structure and camera trajectory estimated by our method are much more closer to the `Ground Truth'. More importantly, our method allows ORB-SLAM to run successfully on the entire KITTI Raw test sequence without significant drifts from the result of `Ground Truth' --- see Fig.~\ref{fig:kitti_all} for typical reconstructed scene structure and camera trajectory. In addition, we quantitatively evaluate the camera trajectory computed by our method and `Ground Truth' on the entire KITTI Raw test sequence using the public EVO toolbox~\footnote{https://michaelgrupp.github.io/evo/}. Tab.~\ref{tab:rmse} presents the median root mean square error (RMSE) of the keyframe trajectory over 5 runs by our method and `Ground Truth'. It can be seen that the error gap between our method and `Ground Truth' is marginal, while our method does not require calibration with a checkerboard pattern.

\begin{figure}[!!t]
	\begin{center}
		\includegraphics[width=0.95\linewidth, trim = 0mm 0mm 80mm 15mm, clip]{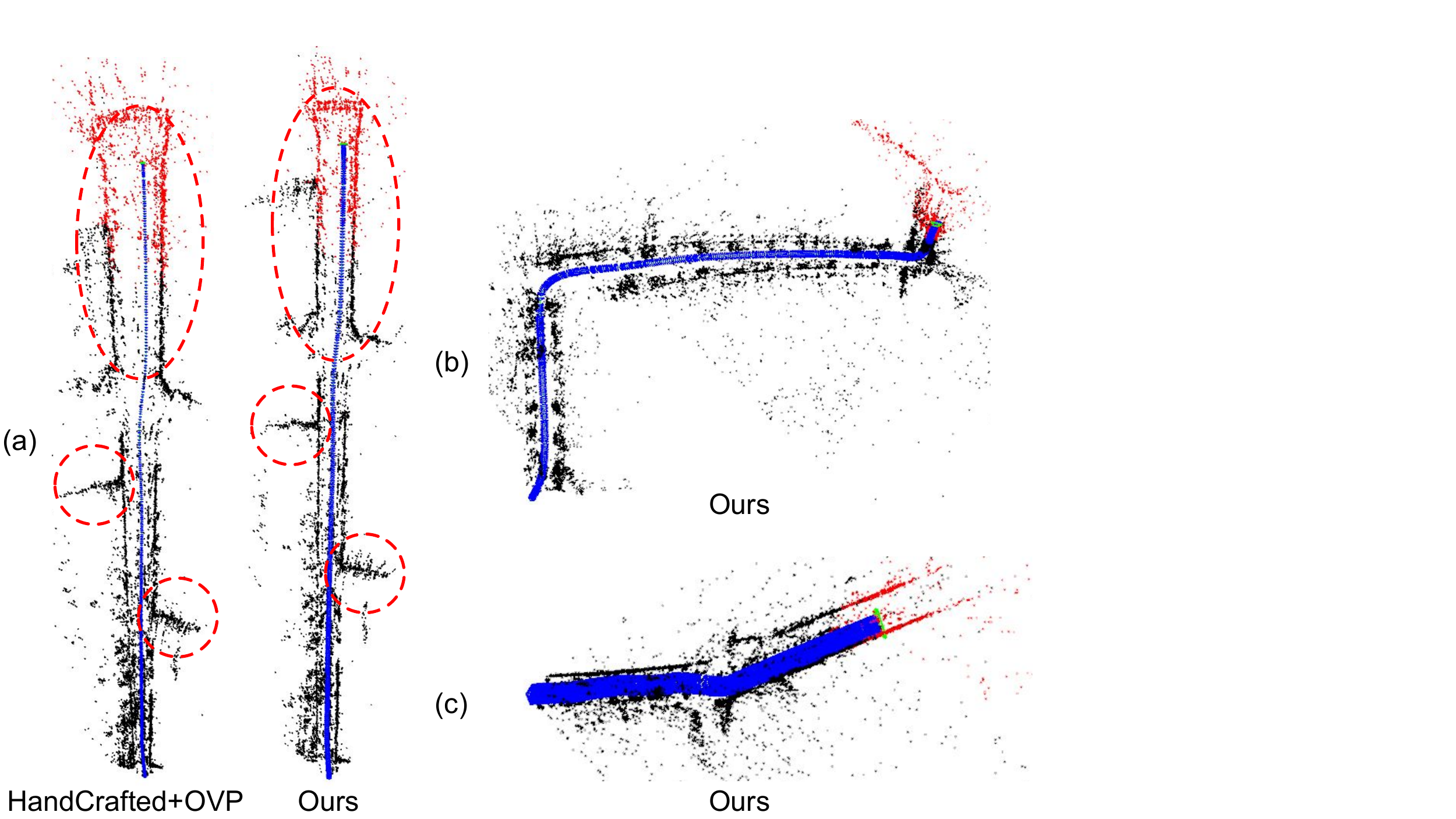}
	\end{center}
	\vspace{-0.4cm}
	\caption{Qualitative results of uncalibrated SLAM on YouTube test sequences. (a) Sequence 1. (b) Sequence 2. (c) Sequence 3.  Red points denote the latest local map.}
	\label{fig:youtube}
\end{figure}

\vspace{0.2cm}
\noindent \textbf{YouTube: }Fig.~\ref{fig:youtube} illustrates qualitative results of our method and `HandCrafted+OVP' on YouTube test sequences. `Ground Truth' is not included since ground truth parameters are not available for these sequences. While our method enables ORB-SLAM to run successfully for all 3 sequences, `HandCrafted+OVP' breaks ORB-SLAM for the last 2 sequences (i.e., ORB-SLAM cannot initialize and hence completely fails). In addition, although ORB-SLAM does not fail in the first sequence for `HandCrafted+OVP', the reconstructed scene structure is more distorted compared to our method. For instance, the parallelism of the two sides of the road is preserved better by our method (see red ellipses). Please see the supplementary video\footnote{https://youtu.be/cfWq9uz2Zac} for more details.

\section{Conclusion}
\label{sec:conclusion}
In this paper, we revisit the camera self-calibration problem which still remains an important open problem in the computer vision community. We first present a theoretical study of the degeneracy in two-view geometric approach for radial distortion self-calibration. A deep learning solution is then proposed and contributes towards the application of SLAM to uncalibrated videos. Our future work includes gaining a better understanding of the network, e.g., using visualization tools such as Grad-CAM~\cite{selvaraju2017grad}. In addition, we want to work towards a more complete system that allows SLAM to succeed in videos with more challenges, e.g. motion blurs and rolling shutter distortion, as well as explore SfM~\cite{zhuang2018baseline}, 3D object localization~\cite{srivastava2019learning} and top-view mapping~\cite{wang2019parametric} in unconstrained scenarios.

\vspace{0.2cm}
\noindent \textbf{Acknowledgements: } Part of this work was done during B. Zhuang's internship at NEC Labs America. This work is also partially supported by the Singapore PSF grant 1521200082 and MOE Tier 1 grant R-252-000-A65-114.

{\small
\bibliographystyle{IEEEtran}
\bibliography{reference}
}

\end{document}